\documentclass[11pt,letterpage]{article}
\usepackage[top=1in,bottom=1in,left=1in,right=1in]{geometry}
\usepackage{natbib}
\usepackage{amsmath,amssymb}
\usepackage{amsfonts}

\usepackage{nameref,hyperref}

\usepackage{graphicx}

\def \boldmu{\boldsymbol{\mu}}
\def \boldA{\boldsymbol{A}}
\def \boldV{\boldsymbol{V}}
\def \boldx{\boldsymbol{x}}

\begin{document}
\title{MRPC: An R package for accurate inference of causal graphs}
\author{Md. Bahadur Badsha$^*$, Evan A Martin, Audrey Qiuyan Fu$^*$ \\ Department of Statistical Science, \\
Institute of Bioinformatics and Evolutionary Studies, \\
Center for Modeling Complex Interactions, University of Idaho \\
$^*$ mdbadsha@uidaho.edu (M.B.B.); audreyf@uidaho.edu (A.Q.F.)}
\date{}
\maketitle


\abstract{
  We present {MRPC}, an R package that learns causal graphs with improved accuracy 
  over existing packages, such as {pcalg} and {bnlearn}.  Our algorithm builds on the powerful PC algorithm, 
  the canonical algorithm in computer science for learning directed acyclic graphs.  The improvement 
  in accuracy results from online control of the false discovery rate (FDR) that reduces false positive edges, a more accurate 
  approach to identifying v-structures (i.e., $T_1 \rightarrow T_2 \leftarrow T_3$), and robust estimation of the correlation 
  matrix among nodes.  For genomic data that contain genotypes and gene expression for each sample, MRPC incorporates 
  the principle of Mendelian randomization to orient the edges.  
  Our package can be applied to continuous and discrete data.
}



\section[Introduction: Causal graph inference in R]{Introduction: Causal graph inference in \texttt{R}} \label{sec:intro}
Graphical models provide a powerful mathematical framework to represent dependence among variables.  Directed edges in a graphical model 
further represent different types of dependencies that may be interpreted as causality \citep{lauritzen1996}.  Directed Acyclic Graphs (DAGs), also 
known as Bayesian networks, are a class of graphical models with only directed edges and no cycles.  However, with real-world problems,  
the true graph contains both directed and undirected edges, or the data may not provide enough information for direction and an edge can be inferred 
only as undirected.  Graphs with both directed and undirected edges are mixed graphs.  We treat undirected edges the same as bidirected ones, as 
not knowing the direction is equivalent to assuming that both directions are equally likely.  We term these mixed graphs {\it causal graphs} (or equivalently, 
causal networks) to emphasize the causal interpretation of the directed edges in the graph.

Multiple DAGs may be equivalent in their likelihoods, and therefore belong to the same Markov equivalent class \citep{richardson1997dag}.  Without additional information, 
inference methods can infer only these Markov equivalent classes.  For example, for a simple graph of three nodes, namely $X$, $Y$ and $Z$, if 
$X$ and $Z$ are conditionally independent given $Y$ (i.e., $X{\perp\!\!\!\perp}Z \; | \; Y$), three Markov equivalent graphs exist:
\begin{align}
X{\perp\!\!\!\perp}Z\; | \;Y: \;\; X\rightarrow Y \rightarrow Z; \;\; X\leftarrow Y \leftarrow Z; \;\; X\leftarrow Y\rightarrow Z.
\end{align}
Without additional information, it is not possible to determine which graph is the truth.

Due to Markov equivalence, existing methods typically can infer only a Markov equivalence class of DAGs that can be uniquely described by completed partially 
directed acyclic graph (CPDAG), which is a mixed graph.  Many methods have been developed, which can be broadly classified as i) constraint-based 
methods, which perform statistical tests of marginal and conditional independence, ii) scored-based methods, which optimizes the search according to a score function; 
and iii) hybrid methods that combine the former 
two approaches \citep{scutari2010.bnlearn}.  The PC algorithm (named after its developers Peter Spirtes and Clark Glymour) is one of the first constraint-based algorithms 
\citep{Spirtes2000}.  Through a series of statistical tests of marginal and conditional independence, this algorithm makes it computationally feasible to infer graphs 
of high dimensions, and has been implemented in open-source software, such as the \texttt{R} package \texttt{pcalg} \citep{Kalisch.eta.2012.pcalg}.
The PC algorithm consists of two main steps:
skeleton identification, which infers an undirected graph from the data;, and edge orientation, which determines the direction of edges in the skeleton.  The \texttt{R} package 
\texttt{bnlearn} implements a collection of graph learning methods from the three classes \citep{scutari2010.bnlearn}.  
In particular, the function \texttt{mmhc()} implements a hybrid algorithm that also consists 
of two steps: learning the neighbors (parent and child nodes) of a node, and finding the graph that is consistent with the data and the neighbors identified from 
the first step \citep{Tsamardinos2006}.

\begin{figure}[t!]
\centering
\includegraphics[scale=0.7]{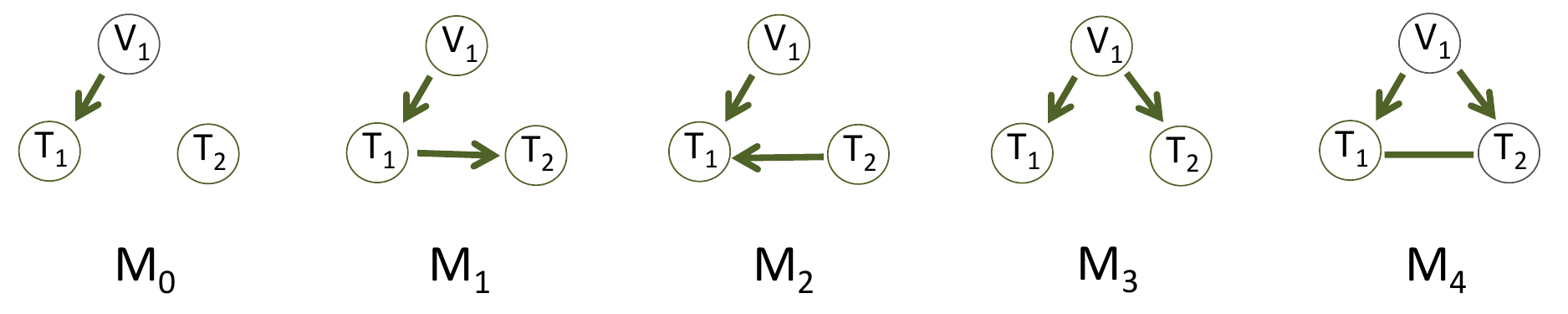}
\caption{Five basic causal graph under the principle of Mendelian randomization.  These graphs all include a genotype node, $V_1$, and two phenotype nodes, $T_1$ and 
$T_2$.  Graphs $M_1$ through $M_4$ are the four scenarios where $T_1$ and $T_2$ are correlated, although due to different causal relationships.  Graph $M_0$ is the 
null graph where $T_1$ and $T_2$ are not causally related or correlated.}
\label{fig:basics}
\end{figure}

The methods described above are designed for generic scenarios.  In genomics there is growing interest in learning causal graphs among 
genes or other biological entities.  Biological constraints, such as the principle of Mendelian randomization (PMR), may be explored for efficient inference of causal graphs 
in biology.  Loosely speaking, the PMR assumes that genotype causes phenotype and not the other way around.  Specifically, people in a natural population may carry 
different genotypes at the same locus on the genome.  Such locus is a genetic variant and can be used as an instrumental variable in causal inference.  
Different genotypes that exist in the population 
can be thought of as natural perturbations randomly assigned to individuals.  Therefore, an association between these genotypes and a phenotype (e.g., expression of a nearby 
gene) can be interpreted as the genetic variant being causal to the phenotype.  This principle has been explored in genetic epidemiology in recent years 
\citep{davey2014mendelian}.  Under the PMR, five causal graphs involving a genetic variant node and two phenotype nodes are possible (Fig~\ref{fig:basics}; 
also see Fig 1 in \citet{Badsha2017}).  These five basic causal 
graphs add constraints to the graph inference process, and allow us to develop the MRPC algorithm that learns a causal graph accurately for hundreds of nodes from 
genomic data \citep{Badsha2017}.

Our algorithm, namely MRPC, is essentially a variant of the PC algorithm that incorporates the PMR.  Similar to other PC algorithms, our method also 
consists of two steps: skeleton identification and edge orientation.  We incorporate the PMR in the second step.  Meanwhile, our MRPC algorithm is not limited to genomic data.
We have developed several improvements over existing methods, and these improvements enable us to obtain more accurate, stable and efficient inference for generic data 
sets compared to \texttt{mmhc} in the \texttt{bnlearn} package and \texttt{pc} in the \texttt{pcalg} package.  In this paper, we explain the details of our MRPC algorithm, 
describe the main functions in the \texttt{R} package \texttt{MRPC}, demonstrate its functionality with examples, and highlight the key improvements over existing methods.  See \cite{Badsha2017} for comparison with other methods based on the PMR and application to genomic data.

\section{Models and software} \label{sec:models}

As PC-based algorithms have demonstrated efficiency in learning causal graphs, we built our algorithm on the \texttt{pc} function implemented in 
the \texttt{R} package \texttt{pcalg}.  Similar to other PC-based algorithm, our MRPC algorithm consists of two steps: learning the graph skeleton, 
and orienting edges in the skeleton.  

The key input of the \texttt{MRPC} function includes the following:
\begin{itemize}
  \item A $N\times M$ data matrix with each row being one of $N$ individuals, and each column one of $M$ features (i.e., a node).  Each entry may be continuous or discrete.
  \item The $M \times M$ correlation matrix calculated for the features in the data matrix.  If the correlation matrix is provided, then the data matrix is ignored.  Otherwise, the correlation matrix is calculated from the data matrix.  \texttt{MRPC()} allows for calculation of the robust correlation matrix for continuous variables.
  \item A desirable overall false discovery rate (FDR).
  \item The number of genetic variants.  This number is 0 if the data do not contain such information.
\end{itemize}

\subsection{Step I: skeleton identification}
In order to learn the graph skeleton, which is an undirected graph, we use the procedure implemented in \texttt{R} function \texttt{pc()} in \texttt{pcalg}, but apply the LOND algorithm 
to control the false discovery rate \citep{javanmard2018.lond}.  The procedure in \texttt{pc()} is standard in all the PC-based algorithms: it starts with a fully connected graph, and 
then conducts a series of statistical tests for pairs of nodes, removing the edge if the test for independence is not rejected.  The statistical tests 
consist of those testing marginal 
independence between two nodes, followed by those of conditional independence tests between two nodes, given one other node, two other nodes, 
and so on.  Each of the tests produces a $p$ value, and if the $p$ value is larger than the threshold, then the two nodes are independent and the edge between 
them is removed.  Following removal of this edge, the two nodes involved will not be tested again in this step.  

The first step involves a sequence of many tests, with the number of tests unknown beforehand.  We have implemented the LOND method in our algorithm to control the 
overall false discovery rate in an online manner \citep{javanmard2018.lond}.  Specifically, each time we need to conduct a test, we use the LOND method to calculate 
a $p$-value threshold for this test.  Depending how many tests have been performed and how many rejections have been made, the $p$-value thresholds tend to be large 
at the beginning and decrease as more tests are performed.

\subsection{Step II: edge orientation}
We design the following steps for edge orientation:
\begin{enumerate}
  \item If the data contain genotype information and there are edges connecting a genotype node to a non-genotype node, then the edge should always point 
  to the non-genotype node, reflecting the assumption in the PMR that genotypes influence phenotypes, but not the other way round.
  \item Next, we search for possible triplets that may form a v-structure (e.g., $X_1\rightarrow X_2 \leftarrow X_3$).  We check conditional test results from step I to 
  see whether $X_1$ and $X_3$ are independent given $X_2$.   If they are, then this is not a v-structure; alternative models for the triplet may be any one of the following three: $X_1 \rightarrow X_2 \rightarrow X_3$, $X_1 \leftarrow X_2 \leftarrow X_3$, and $X_1\leftarrow X_2\rightarrow X_3$.  If this test is not performed in the 
  first step, we conduct it now.
  \item If there are undirected edges after steps (1) and (2), we look for triplets of nodes with at least one directed edge and no more than one undirected edge.
  We check the marginal and conditional test results from step I to determine which of the basic models is consistent with the test results (Fig.~\ref{fig:basics}).  If we can identify 
  such a basic model, we determine the direction of the undirected edge.  We then move on to another candidate triplet.  We go through all undirected edges 
  in this step until all of them have been examined.  It is plausible that some undirected edges cannot be oriented (see examples below), and we leave them 
  as undirected.  Thus, the resulting graph may have both directed and undirected edges, but no directed cycles.
\end{enumerate}


\subsection{Hypothesis testing for marginal and conditional independence}
A variety of tests may be performed 
in this step.  \texttt{pcalg} includes a commonly-used test for continuous values, which is the (partial) correlation test using Fisher's z transformation \citep{Kalisch2007}.  
Briefly, the test statistic between nodes $x$ and $y$ given $S$, a set of other nodes in the graph, is
\begin{align}
T = \sqrt{n-|S|-3} \frac{1}{2} \log \frac{1+\hat{r}_{x,y|S}}{1-\hat{r}_{x,y|S}},
\end{align}
where $n$ is the sample size, $|S|$ the number of nodes in the set $S$, $\hat{r}_{x,y|S}$ the estimated (partial) correlation between $x$ and $y$ given $S$.  
For marginal correlation, $S$ is an empty set, and therefore $|S|=0$ and $\hat{r}_{x,y|S}$ is the sample correlation between $x$ and $y$.  This test statistic 
has a standard normal distribution under the null.

For discrete values, pcalg applies the G$^2$ test.  We use the same function in our \texttt{MRPC} package as well.  Briefly, in a $2\times2$ contingency table for two discrete 
variables $x$ of $k$ levels and $y$ of $l$ levels, we have the observed count $O_{ij}$ and expected count $E_{ij}$, where $i=1,\dots, k$ and $j=1, \dots, l$.  The test statistic 
is 
\begin{align}
G^2 = 2 \sum_{i, j} O_{ij} \log \frac{O_{ij}}{E_{ij}},
\end{align}
which follows a $\chi^2$ distribution with $(k-1)(l-1)$ degrees of freedom.  When testing for conditional independence between $x$ and $y$ given $S$ with $m$ levels, 
the test statistic is
\begin{align}
G^2 = 2 \sum_s \sum_{i, j} O_{ij|s} \log \frac{O_{ij|s}}{E_{ij|s}},
\end{align}
where $O_{ij|s}$ and $E_{ij|s}$ are the observed and expected count for the $i$th level in $x$ and the $j$th level in $y$, given the $s$th level in $S$.  This statistic has 
a $\chi^2$ distribution with $m(k-1)(l-1)$ degrees of freedom under the null.

\subsection{Pearson and robust correlation}
The main input to the \texttt{MRPC()} function is the correlation matrix calculated from the data matrix.  Pearson correlation is typically used.  When outliers are suspected 
in the data with most nodes having continuous measurements, one may provide the data matrix as the input, and the \texttt{MRPC()} function calculates a robust correlation 
matrix internally.  We implemented the robust correlation calculation method in \cite{Badsha2013}. Specifically, for data that are normal or approximately normal, 
we calculate iteratively the robust mean vector $\boldmu$ of length $M$ and the robust covariance matrix $\boldV$ of $M\times M$ until convergence. At the $t+1$st iteration,
\begin{align}
\boldmu_{t+1} = \frac{\sum^n_{i=1} \phi_\beta (\boldx_i; \boldmu_t, \boldV_t) \boldx_i}{\sum^n_{i=1} \phi_\beta (\boldx_i; \boldmu_t, \boldV_t)},   \label{eqn:rcor.mu}
\end{align}
and
\begin{align}
\boldV_{t+1} = \frac{\sum^n_{i=1} \phi_\beta (\boldx_i; \boldmu_t, \boldV_t) (\boldx_i - \boldmu_t) (\boldx_i - \boldmu_t)'}{(1+\beta)^{-1} \sum^n_{i=1} \phi_\beta (\boldx_i; \boldmu_t, \boldV_t)},  \label{eqn:rcor.v}
\end{align}
where
\begin{align}
\phi_\beta (\boldx_i; \boldmu, \boldV) = \exp \bigg[- \frac{\beta}{2} (\boldx_i - \boldmu)' \boldV^{-1} (\boldx_i - \boldmu) \bigg].  \label{eqn:rcor.phi}
\end{align}
In the equations above, $\boldx_i$ is the vector of continuous measurements (e.g., gene expression) in the $i$th sample, $n$ the sample size, and $\beta$ the 
tuning parameter. Equation~\ref{eqn:rcor.phi} downweighs the outliers through $\beta$, which takes values in $[0,1]$.  Larger $\beta$ leads to smaller weights on 
the outliers. When $\beta=0$,  Equation~\ref{eqn:rcor.v} becomes a biased estimator of the variance, with the scalar $1/n$. When the data matrix contains missing 
values, we perform imputation using the R package  \texttt{mice}~\citep{buuren2010mice}. Alternatively, one may impute the data using other appropriate methods, 
and calculate Pearson or robust correlation matrix as the input for \texttt{MRPC()}.

\subsection{Metrics for inference accuracy}
We provide three metrics for assessing the accuracy of the graph inference.
\begin{itemize}
  \item Recall and precision: Recall (i.e., power, or sensitivity) measures how many edges from the true graph a method can recover, whereas precisions 
  (i.e., 1-FDR) measures how many correct edges are recovered in the inferred graph., as follows:
  \begin{align}
  \text{Recall} &= \frac{K_{TP}}{K_{\text{true}}}; \\
  \text{Precision} &= \frac{K_{TP}}{K_{\text{inferred}}},  
  \end{align}  
  where $K_{TP}$ is the count of edges correctly recovered by the method, $K_{\text{true}}$ the number of edges in the true graph, and $K_{\text{inferred}}$ the 
  number of edges in the inferred graph.However, we consider it more important to be able to identify the presence of an edge than to also get the direct correct.  Therefore, 
  we assign 1.0 to an edge with the correct direction and 0.5 to an edge with the wrong direction or no direction.  For example, when the true graph is 
  $V\rightarrow T_1\rightarrow T_2$ with 2 true edges, and the inferred graphs are i) $V \rightarrow T_1 \rightarrow T_2$, and $V\rightarrow T_2$; ii) $V\rightarrow T_1-T_2$; 
  and iii) $V\rightarrow T_1\leftarrow T_2$, the number of correctly identified edges is then 2, 1.5 and 1.5, respectively.  Recall is calculated to be 2/2=100\%, 1.5/2=75\%, 
  and 1.5/2=75\%, respectively, whereas precision is 2/3=67\%, 1.5/2=75\%, and 1.5/2=75\%, respectively.
  \item Adjusted Structural Hamming Distance (aSHD): The SHD, as is implemented in the \texttt{R} package \texttt{pcalg} and \texttt{bnlearn}, counts how many differences 
  exist between two directed 
  graphs. This distance is 1 if an edge exists in one graph but missing in the other, or if the direction of an edge is different in the two graphs. The larger this distance, the 
  more different the two graphs are. Similar to our approach to recall and precision, we adjusted the SHD to reduce the penalty on the wrong direction of an edge to 0.5. 
  \item Number of unique graphs: in our simulation we typically generate multiple data sets under the same true graph.  We are also interested in how many unique graphs 
  a method can infer from these data sets.  For example, when the true graph is $V\rightarrow T_1\rightarrow T_2 \rightarrow T_3$, two inferred graphs may be $V\rightarrow T_1\leftarrow T_2 \rightarrow T_3$ and $V\rightarrow T_1\rightarrow T_2 \leftarrow T_3$.  However, the number of wrongly inferred edges is one in both cases, leading to 
  identical recall, precision and aSHD.  To distinguish these two graphs, we start with the adjacency matrix for the inferred graph, denoted by $\boldA = \{ a_{ij} \}$, where 
  $a_{ij}$ takes on value 1 if there is a directed edge from node $i$ to node $j$, and 0 otherwise. Next we convert this matrix to a binary string, concatenating all the values 
  by row, and treat this binary string as a binary number.  We then convert the binary number to decimal, and subtract the decimal number corresponding to the true graph.  
  This decimal difference from the truth serves as a unique identification number for the graph.  Additionally, this value being 0 means that the inferred graph is identical to 
  the truth.  We count the uniquely inferred graphs in simulation to examine the variation in the graph inference procedure.  For example, the true and inferred graphs 
  mentioned above are converted to binary strings and then decimal values as follows:
  \begin{align}
  \text{Truth:} \;\; V\rightarrow T_1\rightarrow T_2 \rightarrow T_3: \;\; 0100001000010000 = 16912; \label{eqn:seq.true}\\
  \text{Inferred \#1:} \;\; V\rightarrow T_1\leftarrow T_2 \rightarrow T_3: \;\; 0100000001010000 = 16464; \label{eqn:seq.inferred.1}\\
  \text{Inferred \#2:} \;\; V\rightarrow T_1\rightarrow T_2 \leftarrow T_3: \;\; 0100001000000010 = 16898. \label{eqn:seq.inferred.2}
  \end{align} 
The difference between the two inferred graphs and the truth is $16464 - 16912 = -448$ and $16898 - 16912 = -14$, respectively, thus representing two distinct graphs with edges
  wrongly inferred.
\end{itemize}

\subsection{Simulating continuous and discrete data}
When simulating data under a true graph, we generated data first for the nodes without parents from a marginal distribution, and then for other nodes from 
a conditional distribution. When genetic variants are present in the true graph, they are nodes without parents.  We assume that each genetic variant node is 
a biallelic single nucleotide polymorphism (SNP); that is the variant has two alleles (denoted by 0 and 1) in the population.  Since an individual receives an allele 
from their mother and an allele from their father, the genotype at this variant may be 0 for both alleles, 0 for one allele and 1 for the other, or 1 for both alleles.  
If we count the number of allele 1 in this individual, then the genotype may take on one of three values: 0, 1, and 2.  Let $q$ be the probability of allele 1 in 
the population.  Assume that the probability of one allele is not affected by that of the other allele in the same individual (i.e., the genotypes are in Hardy-Weinberg 
equilibrium).  Then the genotype of the node follows a multinomial distribution:
\begin{align}
\Pr (V=0)= (1-q)^2; \;\; \Pr (V=1)=2q(1-q); \;\; \Pr(V=2)=q^2.
\end{align}
Other types of nodes in the graph that are not genetic variants are denoted by $T$.  Denote the $j$th non-genotype node by $T_j$ and the set of its parent nodes 
by $P$, which may be empty, or may include genotype nodes or other non-genotype nodes. We assume that the values of $T_j$ follows a normal distribution
\begin{align}
T_j \sim N(\gamma_0+\sum_{k\in P} \gamma_k V_k + \sum_{l \in P} \gamma_l T_l, \sigma_j^2 ).
\end{align}
The variance may be different for different nodes. For simplicity, we use the same value for all the nodes.

We treat undirected edges as bidirected edges and interpret such an edge as an average of the two directions with equal weights. For example, for graph 
$M_4$ in Fig.~\ref{fig:basics}, we consider that the undirected edge is a mixture of the two possible directions.  Therefore, we generate data for $T_1 \rightarrow T_2$:
\begin{align}
T_1 \sim N(\gamma_0  +\gamma_1 V, \sigma_1^2 ); \;\; T_2 \sim N(\gamma_0  + \gamma_1 V + \gamma_2 T_1, \sigma_2^2),
\end{align}
and separately for $T_1 \leftarrow T_2$:  
\begin{align}
T_1 \sim N(\gamma_0  + \gamma_1 V + \gamma_2 T_2, \sigma_1^2);  T_2 \sim N(\gamma_0  + \gamma_1 V, \sigma_2^2).
\end{align}
We then randomly choose a pair of values with 50:50 probability for each sample.
For simplicity in simulation, we set $\gamma_0=0$ and all the other $\gamma$s to take the same value, which reflects the strength of the association signal.

The procedure above describes how continuous data are generated.  For discrete data, we use the same procedure and then discretize the generated continuous 
values into categories. 

\subsection{The LOND method for online FDR control}
The LOND algorithm controls FDR in an online manner \citep{javanmard2018.lond}.  This is particularly useful when we do not know the number of tests beforehand 
in learning the causal graph.  Specifically, consider a sequence of null hypotheses (marginal or conditional independence between two nodes), 
denoted as $H^i_0$, with corresponding p-value, $p_i$, $i=1,\dots, m$. 
The LOND algorithm aims to determine a sequence of significance level $\alpha_i$, such that the decision for the $i$th test is 
\begin{align}
R_i = \begin{cases}
	 1, & \text{if $p_i \leq \alpha_i$ (reject $H^i_0$)} \\
	 0, & \text{if $p_i > \alpha_i$ (accept $H^i_0$)}
	 \end{cases}.
\end{align}
The number of rejections over the first $m$ tests is then
\begin{align}
 D(m) = \sum^m_{i=1} R_i.
\end{align}
For the overall FDR to be $\delta$, we use a series of nonnegative values $\delta_i$, and set the significance level $\alpha_i$ as 
\begin{align}
\alpha_i = \delta_i [D(i-1)+1],                            		 			                    
\end{align}
where the FDR for the $i$th test is 
\begin{align}
\delta_i= \frac{c}{i^a},
\end{align}
with 
\begin{align}
\sum_{i=1}^\infty \delta_i = \delta,
\end{align}
for integer $a>1$ and constant $c$. The default value for $a$ is set to 2 in MRPC. At an FDR of 0.05 and $a=2$, we have
\begin{align}
\sum_{i=1}^\infty \delta_i = \sum_{i=1}^\infty \frac{c}{i^2} = c \sum_{i=1}^\infty \frac{1}{i^2} = \frac{c\pi^2}{6} = 0.05.                           				        
\end{align}
Then
\begin{align}
c = (6 \times 0.05)/ \pi^2 = 0.0304.                                                        				     
\end{align}
The larger $a$ is, the more conservative the LOND method, which means that fewer rejections will be made.  We therefore set $a=2$ throughout 
simulation and real data analyses.

\section{Main functions of the MRPC package}
\subsection{Causal graph inference}
\begin{itemize}
  \item \texttt{MRPC()}: this function is central to our package and implements the causal graph inference algorithm described in the previous section.  The required arguments 
  are \texttt{data},  \verb/suffStat/, \verb/GV/, \verb/FDR/, \verb/indepTest/, and \verb/labels/.  \verb/data/ is the data matrix that may contain discrete or continuous values, and 
  may or may not be genomic data. \texttt{suffStat} is a list of two elements: the (robust) correlation matrix calculated from the data matrix, and the sample size. \texttt{GV} is the 
  number of genetic variants, and should be set to 0 if the data is not genomic or does not have genotype information.  \texttt{FDR} is the desirable false discovery rate, a value 
  between 0 and 1.  \texttt{indepTest} specifies the statistical test for marginal and conditional independence; typically, this argument is set to \texttt{gaussCItest} for continuous 
  values, or to \verb/disCItest/ for discrete data.  Lastly, \texttt{labels} is the names of the nodes used for visualization.
  \item \texttt{ModiSkeleton()}: this is a modified version of the skeleton function in the pcalg package, and is called within the MRPC function.  
  Similar to skeleton(), this function also learns the skeleton of a causal graph.
  However, we have incorporated the LOND algorithm into this function such that the overall FDR is controlled.
  \item \texttt{EdgeOrientation()}: this function is called by \verb/MRPC()/ and performs Step II of the MRPC algorithm.  
  \item \texttt{RobustCor()}: this function calculates the robust correlation matrix for a data matrix that does not contain genotype information and have continuous values 
  in all columns (nodes), or that contains genotype information and have continuous values in the phenotype columns.  This function may be used when the above 
  data matrix may contain outliers.  It returns a correlation matrix that may be used as an argument for \texttt{MRPC()}.
\end{itemize}
\subsection{Simulation}
\begin{itemize}
  \item \texttt{SimulatedData()}: this function generates genomic data that contains genetic variants.  
\end{itemize}

\subsection{Visualization}
\begin{itemize}
  \item \texttt{plot()}: this function inherits the functionality of the \texttt{plot} function in the \texttt{pcalg} package.  Therefore, if the graph does not display correctly, the user may 
  need to run \texttt{library (pcalg)} to ensure that the \texttt{pcalg} package is loaded into \texttt{R}.
  \item \texttt{DendroModuleGraph()}: this function is modified from the function with the name in the \texttt{R} package \texttt{WGCNA} \citep{langfelder2008wgcna}.  
  Similar to the \texttt{WGCNA} function, 
  this function also generates a dendrogram of all the nodes in the inferred causal graph based on the distance (i.e., the number of edges) 
  between two nodes, cluster the nodes into modules according to the dendrogram and the minimum module size (i.e., the number of nodes in a module), and draws 
  the inferred causal graph, where nodes of different modules are colored differently.  For genomic data, this function draws the genotype nodes in filled triangles and phenotype 
  nodes in filled circles.
\end{itemize}

\subsection{Assessment of inferred graphs}
\begin{itemize}
  \item \texttt{Recall\_Precision()}: this function compares an inferred graph to a true one, counts the number of true positives and false positives in the inferred graph, and calculates 
  recall and precision.
  \item \texttt{aSHD()}: this function calculates the (adjusted) SHD between two graphs.
  \item \texttt{seqDiff()}: this function converts two graphs into binary strings and then a decimal number, and calculates the difference between these two numbers.  See the example 
  in Eqns~\ref{eqn:seq.true} -- \ref{eqn:seq.inferred.2}.
\end{itemize}

\section{Illustrations} \label{sec:illustrations}
\subsection{Continuous data with and without genetic information}
\begin{figure}[t!]
\centering
\includegraphics[scale=0.8]{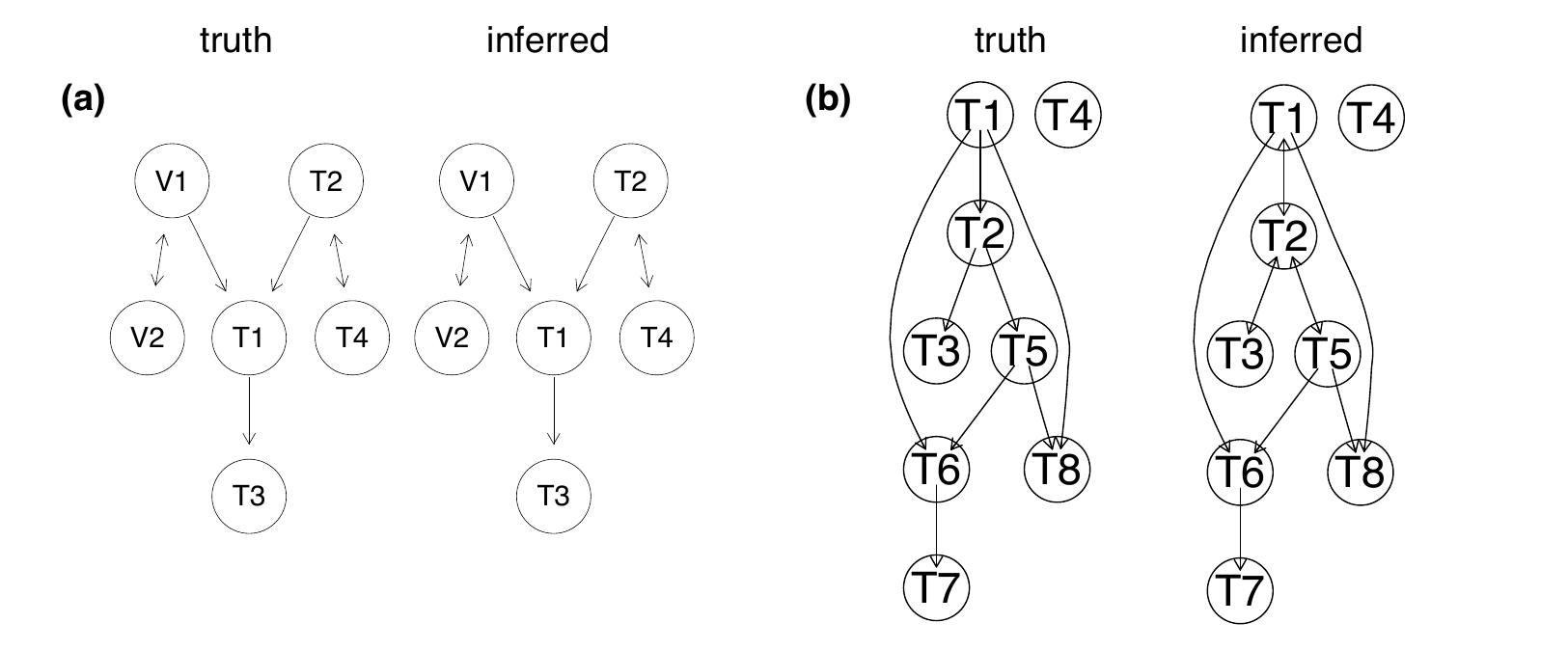}
\caption{Illustration of the \texttt{MRPC} function on continuous data with and without genetic information. (a) The data contain two genotype nodes ($V_1$ and $V_2$).  
\texttt{MRPC} recovers the truth perfectly.  (b) The data are not genomic data.  \texttt{MRPC} does not have enough information to determine the direction of edges 
$T_1-T_2$, $T_2-T_3$, and $T_2-T_5$.}
\label{fig:illustration.cont}
\end{figure}

The first example is a genomic data set with two genotype nodes, $V_1$ and $V_2$, and 4 phenotype nodes, $T_1$ through $T_4$ (Fig.~\ref{fig:illustration.cont}a).  The genotype nodes are discrete, whereas the phenotype nodes are continuous.
%
\begin{verbatim}
R> data ("ExampleMRPC") 
R> n <- nrow (ExampleMRPC$simple$cont$withGV$data) 
R> V <- colnames(ExampleMRPC$simple$cont$withGV$data) 
R> Rcor_R <- RobustCor(ExampleMRPC$simple$cont$withGV$data, Beta=0.005)
R> suffStat_R <- list(C = Rcor_R$RR, n = n)
R> data.mrpc.cont.withGV <- MRPC(data=ExampleMRPC$simple$cont$withGV$data, 
 +	suffStat=suffStat_R, GV=2, FDR=0.05, indepTest='gaussCItest', 
 +	labels=V, verbose = TRUE)

R> par (mfrow=c(1,2))
R> plot (ExampleMRPC$simple$cont$withGV$graph, main="truth") 
R> plot (data.mrpc.cont.withGV$graph, main="inferred")   
\end{verbatim}
%

The second example is the data set from the \texttt{pcalg} package, in which none of the nodes are genotype nodes, and all the nodes take on continuous values (Fig.~\ref{fig:illustration.cont}b).  The command lines for generating the output are similar to those above, with \texttt{ExampleMRPC\$simple\$cont\$withGV\$data} being 
replaced by \texttt{ExampleMRPC\$simple\$cont\$withoutGV\$data}, and \texttt{GV} is set to 0 when calling \texttt{MRPC()}.

\subsection{Discrete data with and without genetic information}
\begin{figure}[t!]
\centering
\includegraphics[scale=0.8]{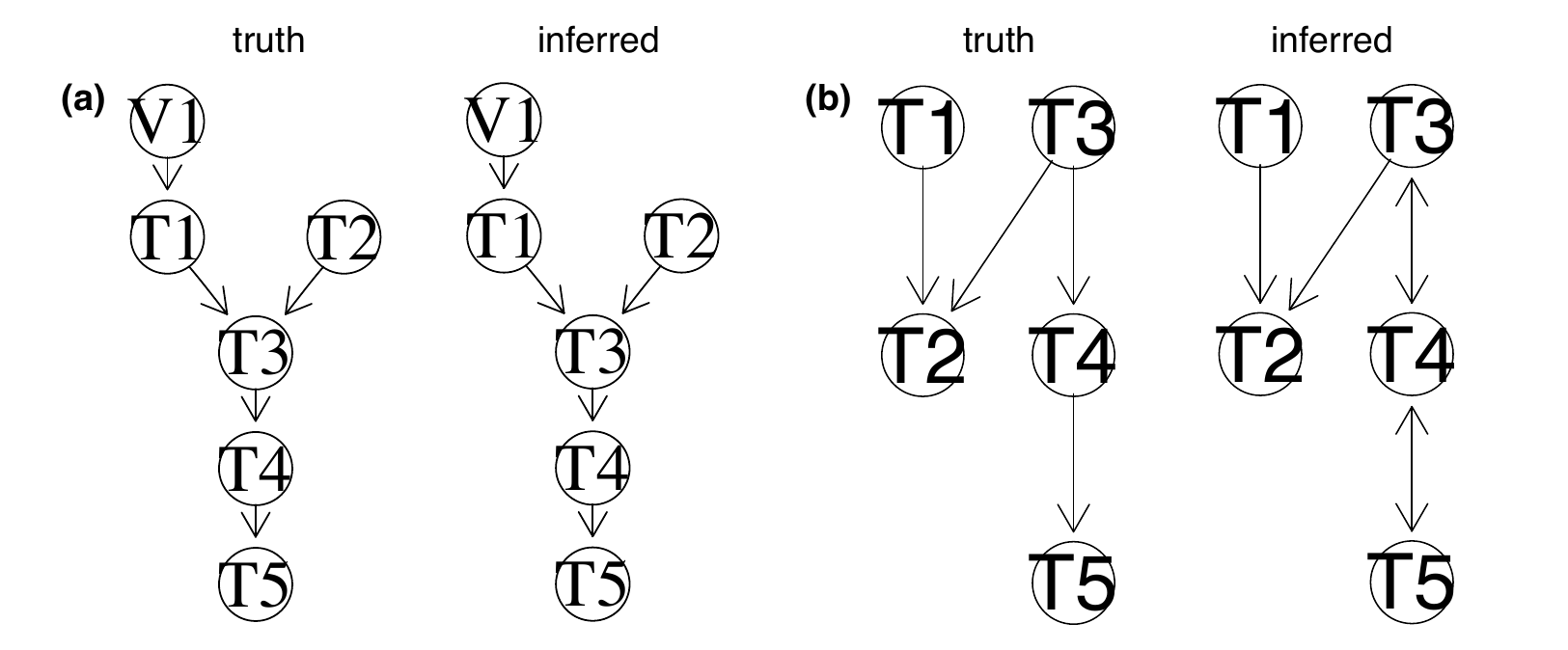}
\caption{Illustration of the \texttt{MRPC} function on discrete data with and without genetic information. (a) The data contain one genetic variant.  \text{MRPC} recovers 
the truth perfectly.  (b) The data are not genomic data.  \texttt{MRPC} does not have enough information to determine the direction of edges $T_3-T_4$ and $T_4-T_5$.}
\label{fig:illustration.disc}
\end{figure}

The first example here is a genomic data set with one genotype node and five phenotype nodes (Fig.~\ref{fig:illustration.disc}a).  All the nodes have discrete values.
%
\begin{verbatim}
R> data ("ExampleMRPC") 
R> n <- nrow (ExampleMRPC$simple$disc$withGV$data) 
R> V <- colnames(ExampleMRPC$simple$disc$withGV$data)  
R> Rcor_R <- RobustCor(ExampleMRPC$simple$disc$withGV$data, Beta=0.005)
R> suffStat_R <- list(C = Rcor_R$RR, n = n)
R> data.mrpc.disc.withGV <- MRPC(data=ExampleMRPC$simple$disc$withGV$data, 
 +	suffStat=suffStat_R, GV=1, FDR=0.05, indepTest='gaussCItest', 
 +	labels=V, verbose = TRUE)

R> par (mfrow=c(1,2))
R> plot (ExampleMRPC$simple$disc$withGV$graph, main="truth") 
R> plot (data.mrpc.disc.withGV$graph, main="inferred")  
\end{verbatim}
%

The second example is a generic data set with five nodes of discrete values, where no nodes are genetic variants (Fig.~\ref{fig:illustration.disc}b).  The command lines for 
generating the output are similar to those above, with \texttt{ExampleMRPC\$simple\$disc\$withGV\$data} being replaced by \\
\texttt{ExampleMRPC\$simple\$disc\$withoutGV\$data}, and with \texttt{GV} set to 0 when calling \texttt{MRPC()}.

\subsection{Simulation and performance assessment}
We include a small example to demonstrate how simulation is performed and how we use functions implemented in \texttt{MRPC} to compare the inferred graph to the truth.  
The user may run the following command line to see the demo:
%
\begin{verbatim}
R> simu <- SimulationDemo(N=1000, model="truth1", signal=1.0, n_data=2, 
 +	n_nodeordering=6)
R> apply (simu, 2, unique)
\end{verbatim}
%
In this demo, ``truth1" represents the graph $V_1 \rightarrow T_1 \rightarrow T_2 \rightarrow T_3$ (also see Figure~\ref{fig:nodeordering}a), which gives rise to six permutations of the three $T$ nodes (we do not permute all the nodes, as the $V$ nodes, which correspond to the genetic variants, need to be at the beginning of the node list): $\{T_1, T_2, T_3\}$, $\{T_1, T_3, T_2\}$, $\{T_2, T_1, T_3\}$, $\{T_2, T_3, T_1\}$, $\{T_3, T_1, T_2\}$, $\{T_3, T_2, T_1\}$.  For each node list, the demo simulates two independent data sets with 1000 observations, applies \texttt{MRPC}, \texttt{mmhc}, and \texttt{pc} to each data set, 
and calculates the difference between the inferred graph and the truth, after converting each graph into a binary string and then a decimal number.  The output is a matrix of 2 rows and $6\times 3=18$ columns.  The second line above lists 
the unique numbers (representing uniquely inferred graphs) for each permutation inferred by each method.

\subsection{Visualizing a complex graph}
\begin{figure}[t!]
\centering
\includegraphics[scale=1.0]{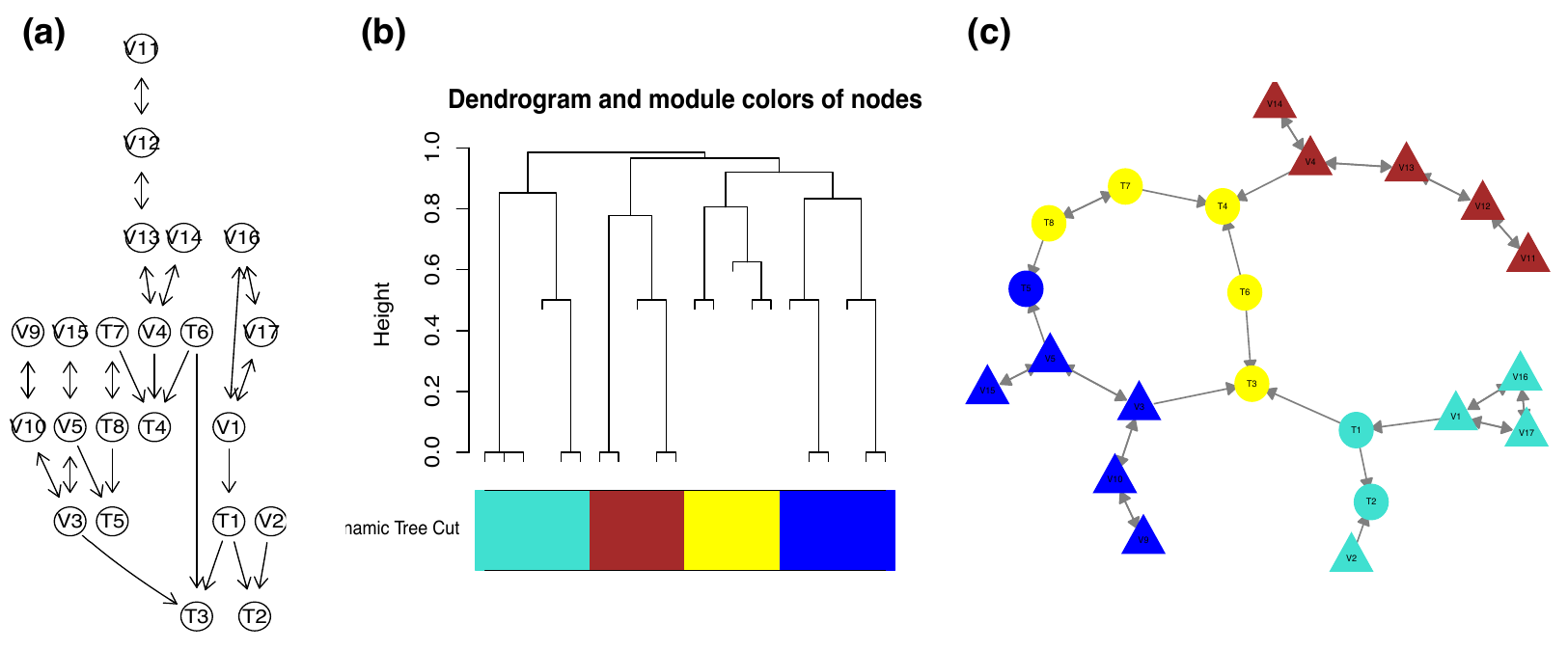}
\caption{Using the \texttt{MRPC} package to visualize complex graphs. (a) A complex graph produced by the \texttt{plot} function shows 14 genetic variants and 
8 phenotype nodes.  (b) The dendrogram of the graph with four modules identified as the minimum module size is set to 5.  (c) The graph with nodes colored 
in the same way as the module they belong to.}
\label{fig:illustration.visual}
\end{figure}

For complex graphs with many nodes, it may help interpretation to cluster the nodes into modules.  Below, we demonstrate this functionality on a 
complex graph of 22 nodes.  This graph may be a true graph, or an inferred one from running the \texttt{MRPC} function.  Figure~\ref{fig:illustration.visual} shows the graph without 
clustering, the dendrogram of the nodes, and the graph with clustering.
%
\begin{verbatim}
R> plot(ExampleMRPC$complex$cont$withGV$graph)

R> Adj_directed <- as(ExampleMRPC$complex$cont$withGV$graph,"matrix")	
R> DendroModuleGraph(Adj_directed, minModuleSize = 5, GV=14)	
\end{verbatim}
%


\section{Key improvements over existing methods}
\subsection{Online FDR control}
Existing methods for graph inference, such as methods implemented in packages \texttt{pcalg} and \texttt{bnlearn}, use only the type I error rate for individual tests and 
do not control for multiple testing.  A key improvement in \texttt{MRPC} over these methods is the incorporation of the LOND method for controlling the false discovery 
rate in an online manner.  As the marginal and conditional independence tests are performed sequentially, the LOND method calculates a desirable type I error rate 
for each test, using the number of tests that have been performed and the number of rejections so far, with the aim of controlling the overall FDR at a pre-determined 
level.

\subsection{V-structure identification}
For any triplet with two edges (i.e., $X-Y-Z$), only two relationships may exist: conditional independence 
$(X{\perp\!\!\!\perp}Z) | Y$, and conditional dependence $(X{\not\perp\!\!\!\perp}Z)|Y$ (we add the parentheses for clarity; these parentheses are typically not used 
in standard notation). 
 Three equivalent graphs exist for conditional independence:
\begin{align}
(X{\perp\!\!\!\perp}Z)|Y: \;\; X\rightarrow Y \rightarrow Z; \;\; X\leftarrow Y \leftarrow Z; \;\; X\leftarrow Y \rightarrow Z.
\end{align}
As mentioned in the Introduction, these three graphs are Markov equivalent and cannot be distinguished.  Only one graph is possible for conditional dependence:
\begin{align}
(X{\not\perp\!\!\!\perp}Z)|Y: \;\; X\rightarrow Y \leftarrow Z,
\end{align}
which is the v-structure.  As the only unambiguous graph one may infer for the same skeleton, it is therefore critical to correctly identify all the v-structures in the data.

\begin{figure}[t!]
\centering
\includegraphics[scale=0.8]{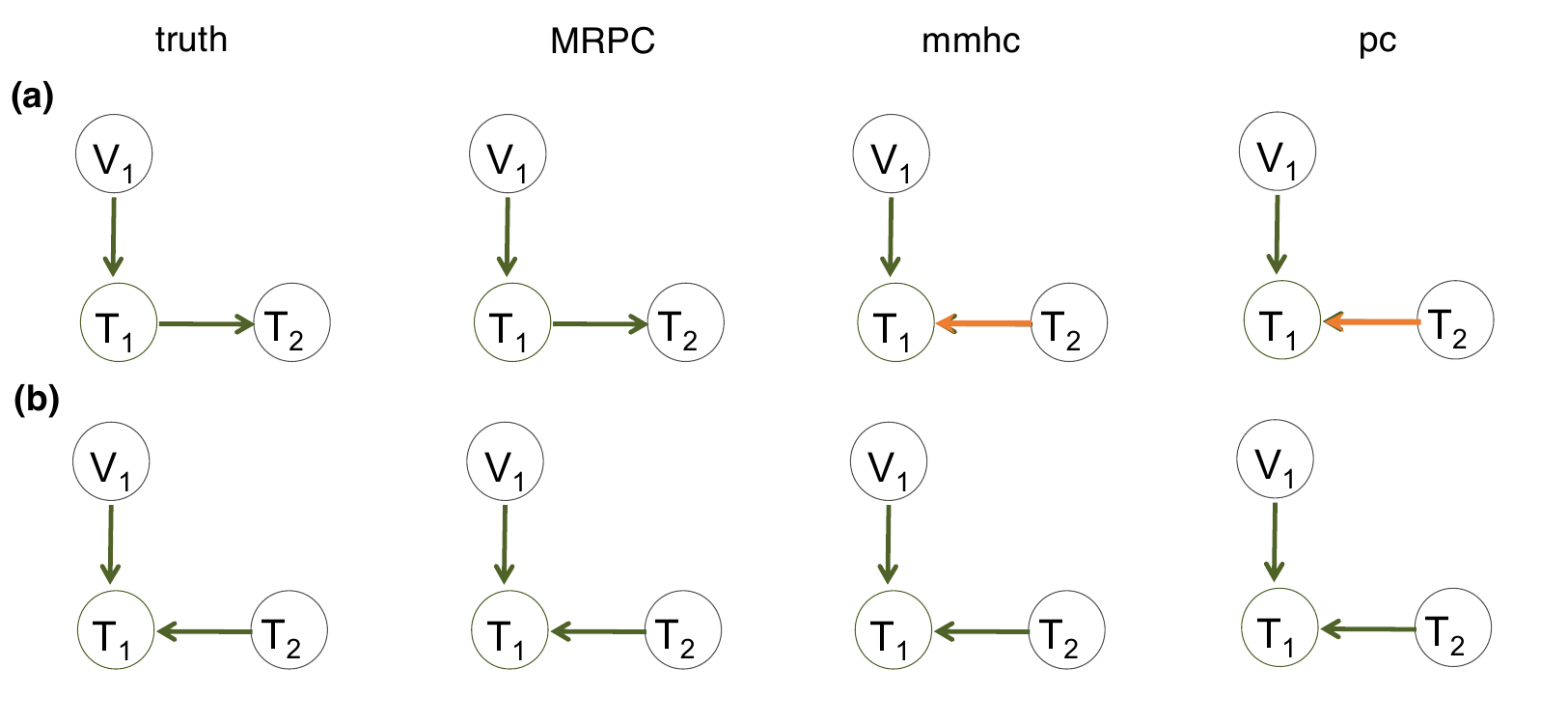}
\caption{V-structure identification among \texttt{MRPC}, \texttt{mmhc} and \texttt{pc}. (a) The true graph does not have a v-structure.  (b) The true graph has a v-structure.}
\label{fig:vstructure}
\end{figure}

However, both \texttt{pc} and \texttt{mmhc} may wrongly identify v-structures when there is not one (Figure~\ref{fig:vstructure}).  
With \texttt{pc}, the false positive is due to incorrect 
interpretation of the lack of the edge $X-Z$.  Specifically, when testing for marginal independence in the first step of skeleton identification, the null hypothesis of 
$X$ and $Z$ being independent is likely not rejected, leading to the removal of the edge $X-Z$.  Once an edge is removed in the inference, the two nodes are not 
going to be tested again.  This means that there is no evidence for or against $(X{\not\perp\!\!\!\perp}Z)|Y$.  However, \texttt{pc} interprets this lack of evidence as 
support for $(X{\not\perp\!\!\!\perp}Z)|Y$ and claims a v-structure.  As a result, \texttt{pc} usually identifies v-structures when they are present, but also falsely calls 
v-structures when there are none.  It is unclear why \texttt{mmhc} also falsely identifies v-structures.  
With \texttt{MRPC}, once we identify a candidate triplet of this skeleton, we examine whether
the conditional independence test between $X$ and $Z$ given $Y$ has been performed before, and if not, conduct this test to determine whether the data supports 
a v-structure.

\subsection{Node-ordering independence}
\begin{figure}[t!]
\centering
\includegraphics[scale=0.8]{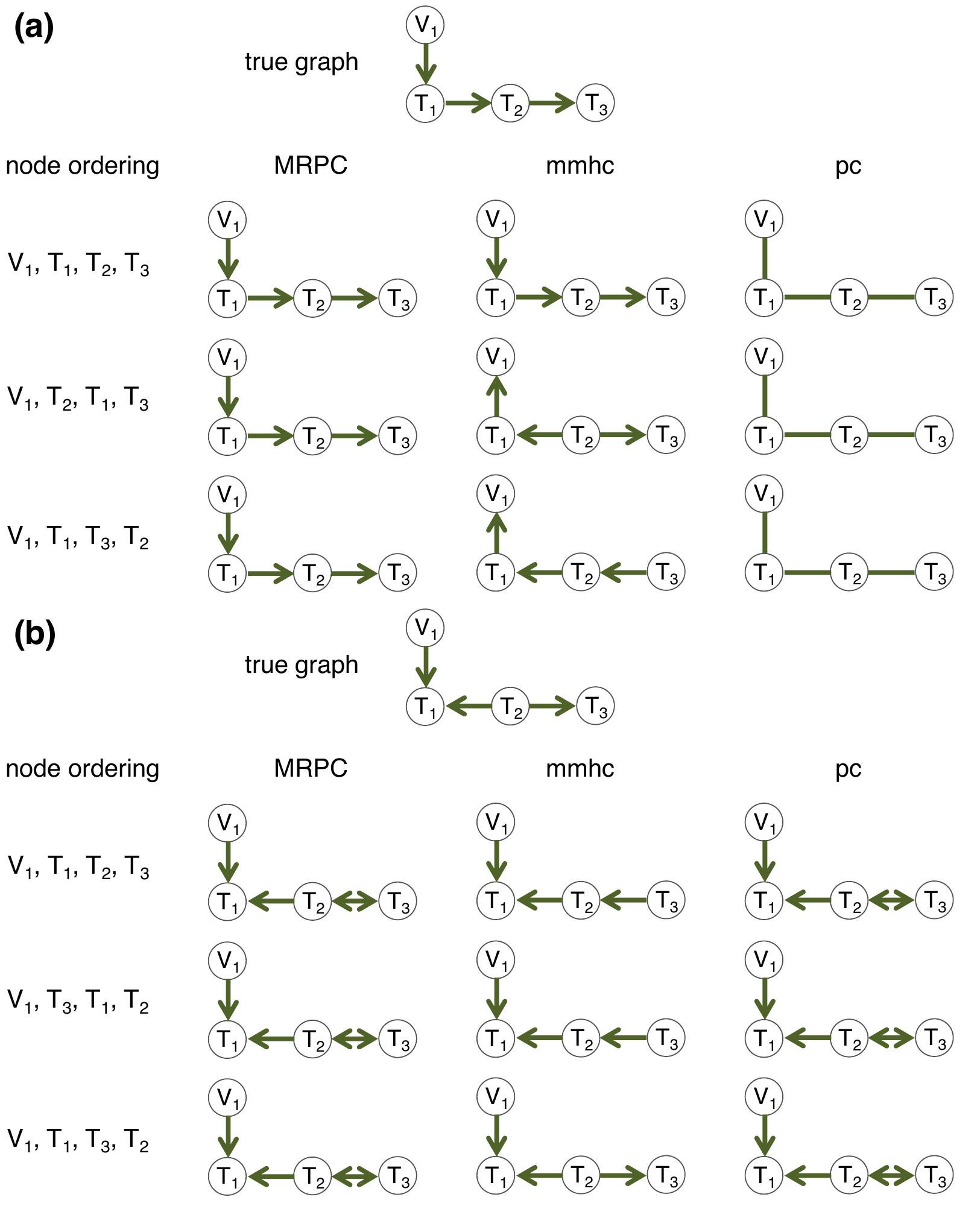}
\caption{Comparison of \texttt{MRPC}, \texttt{mmhc} and \texttt{pc} on node-ordering independence. (a) The true graph does not contain a v-structure.  
(b) The true graph contains a v-structure.}
\label{fig:nodeordering}
\end{figure}

It is desirable to learn the same graph even after the nodes in the data set are permuted.  Our MRPC algorithm demonstrates stability across different simulations, 
better than \texttt{pc}, which implements a node-ordering independent algorithm, and much better than \texttt{mmhc}.  For example, for each of the two graphs in Figure~\ref{fig:nodeordering}, 
we simulated a single data set and permuted the nodes (columns).  Our method processes the genotype nodes ($V_1$) and the phenotype nodes ($T_i$) separately, 
and therefore we permuted only the $T_i$s.  Three $T_i$s have three different permutations.  We then ran \texttt{MRPC}, \texttt{mmhc} and \texttt{pc} on each data set 
to infer the causal graph.  Whereas \texttt{MRPC} and \texttt{pc} produced the same graph on three permutations, \texttt{mmhc} generated a different graph for each 
permutation in 
Figure~\ref{fig:nodeordering}a and two graphs for three permutations in Figure~\ref{fig:nodeordering}b.

We next generated 200 data sets from each of the graphs in Figure~\ref{fig:nodeordering}.  For each data set, we permuted the $T$ nodes to derive six data sets.  
We then applied \texttt{MRPC}, \texttt{mmhc} and \texttt{pc} to all the data sets, counted the number of unique graphs among the six inferred graphs for each data set, and summarized 
the interquartile range, median and max across all 200 data sets for each true graph (Table~\ref{tab:nodeordering}).  
\texttt{mmhc} appears to be more unstable compared to the other two methods. Note that we focus on the variation in inferred graphs, not the accuracy.  Although \texttt{pc} is 
stable, it may not infer the graph correctly (Figure~\ref{fig:nodeordering} (a)).

\begin{table}[t!]
\centering
\begin{tabular}{l | r | r | r | r}
\hline
\hline
 & \multicolumn{4}{c}{Graph (a) in Fig~\ref{fig:nodeordering}} \\
\hline
 & 1st Quartile & Median & 3rd Quartile & Max \\
 \texttt{MRPC} & 1.00  &  1.00  &  1.00  &  2.00\\
 \texttt{mmhc} & 2.00 &   3.00  &  3.00  &  4.00\\
 \texttt{pc} & 1.00  &  1.00  &  1.00  &  2.00\\
\hline
 & \multicolumn{4}{c}{Graph (b) in Fig~\ref{fig:nodeordering}} \\
\hline
 & 1st Quartile & Median & 3rd Quartile & Max \\
 \texttt{MRPC} & 1.00  &  1.00  &  1.00  &  2.00\\
 \texttt{mmhc} & 2.00 &   2.00 &  2.00 &   3.00 \\
 \texttt{pc} & 1.00  &     1.00   &    1.00    &  1.00\\
\hline
\hline
\end{tabular}
\caption{\label{tab:nodeordering} Counts of uniquely inferred graphs with node permutation.}
\end{table}

\subsection{Robustness in the presence of outliers}
\begin{figure}[t!]
\centering
\includegraphics[scale=0.8]{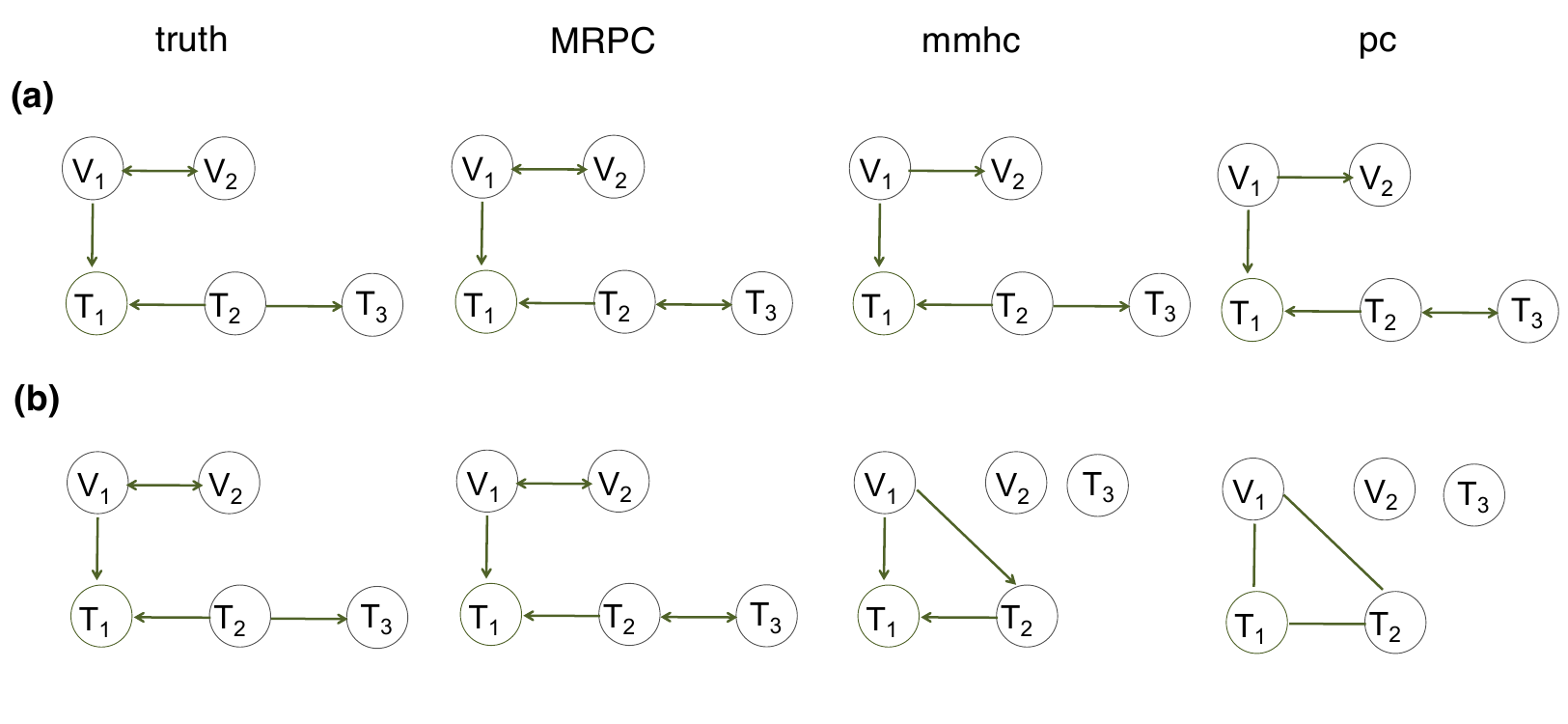}
\caption{Impact of outliers on inference by \texttt{MRPC}, \texttt{mmhc} and \texttt{pc}. (a) The true graph does not contain outliers.  (b) The true graph has 10 outliers among the 1000 observations.}
\label{fig:outliers}
\end{figure}

When the data may contain outliers, our \texttt{MRPC} package allows for calculation of a robust correlation matrix from the data matrix.   Each sample in the data matrix is assigned a 
weight in the correlation calculation.  Outliers tend to receive a weight near 0, thus their contribution to correlation is reduced.  Neither \texttt{pc} or \texttt{mmhc} deals with outliers.  
As our simulation results show, \texttt{MRPC} infers the same graph with or without the presence of outliers, and the inferred graph is close to the truth.  
By contrast, \texttt{pc} and \texttt{mmhc} 
wrongly place edges where they do not belong (Figure~\ref{fig:outliers}).

Our current implementation of the robust 
correlation calculation is limited to continuous data for all the columns if there is no genotype information, and for the phenotype columns if there is genotype information.





\section{Summary and discussion} \label{sec:summary}
Here, we introduce our \texttt{R} package \texttt{MRPC} that implements our novel algorithm for causal graph inference.  Our algorithm 
builds on existing PC algorithms and incorporates the principle of Mendelian randomization for genomic data when genotype and molecular phenotype data 
are both available at the individual level.
Our algorithm also controls the overall FDR for the inferred graph, improves the v-structure identification, reduces dependence on the node ordering, 
and deals with outliers in the data.  These improvements are not limited to genomic data.  Therefore, our MRPC algorithm is a 
much improved algorithm for causal graph learning for generic data.  Our \texttt{MRPC} package contains the main function \texttt{MRPC()} for causal graph learning, as well as functions for simulating genomic and non-genomic 
data from a wide range of graphs, for visualizing the graphs, and for calculating several metrics for performance assessment.

Through simulation, we demonstrated that our method is stable, accurate and efficient on relatively small graphs.  However, due to the online FDR control 
implemented in our package, the inference can be slow when the number of nodes reaches thousands.  We have yet to develop more efficient search algorithms 
while retaining accuracy and stability.  



\section*{Computational details}

The results in this paper were obtained using
\texttt{R}~3.4.1 with the
\texttt{MASS}~7.3.47 package. The \texttt{MRPC} package is available at \url{https://github.com/audreyqyfu/mrpc}.
\texttt{R} itself
and all packages used are available from the Comprehensive
\texttt{R} Archive Network (CRAN) at
\url{https://CRAN.R-project.org/}.

\section*{Acknowledgments}

This research is supported by NIH/NHGRI R00HG007368 (to A.Q.F.) and partially by the NIH/NIGMS grant P20GM104420 to the 
Center for Modeling Complex Interactions at the University of Idaho.


\bibliography{refs_arxiv}


\newpage


\end{document}